\documentclass[letterpaper]{article} 
\usepackage{aaai2026}  
\usepackage{times}  
\usepackage{helvet}  
\usepackage{courier}  
\usepackage[hyphens]{url}  
\usepackage{graphicx} 
\usepackage{natbib}  
\usepackage{caption} 
\usepackage{algorithm}
\usepackage{algorithmic}
\usepackage{amsmath}
\usepackage{booktabs}
\usepackage{siunitx}
\usepackage{multirow}
\usepackage{graphicx}
\usepackage{colortbl}
\usepackage{graphicx}
\usepackage{subcaption}  
\usepackage[most]{tcolorbox}
\usepackage{times}
\usepackage{helvet}
\usepackage{courier}
\usepackage{url}

\usepackage{newfloat}
\usepackage{listings}
\DeclareCaptionStyle{ruled}{labelfont=normalfont,labelsep=colon,strut=off} 
\floatstyle{ruled}
\newfloat{listing}{tb}{lst}{}
\floatname{listing}{Listing}
\title{PolyLingua: Margin-based Inter-class Transformer for Robust Cross-domain Language Detection}
\author{Ali Lotfi Rezaabad, Bikram Khanal, Shashwat Chaurasia, Lu Zeng, Dezhi Hong, Hossein Bashashati, Thomas Butler, Megan Ganji
}

\affiliations{
    Alexa, Amazon\\
    \{alotfi,ganjmegj\}@amazon.com
}

\usepackage{bibentry}

\begin{document}

\maketitle

\begin{abstract}
Language identification is a crucial first step in multilingual systems such as chatbots and virtual assistants, enabling linguistically and culturally accurate user experiences. Errors at this stage can cascade into downstream failures, setting a high bar for accuracy. Yet, existing language identification tools struggle with key cases--such as music requests where the song title and user language differ. Open-source tools like LangDetect, FastText are fast but less accurate, while large language models, though effective, are often too costly for low-latency or low-resource settings. We introduce PolyLingua, a lightweight Transformer-based model for in-domain language detection and fine-grained language classification. It employs a two-level contrastive learning framework combining instance-level separation and class-level alignment with adaptive margins, yielding compact and well-separated embeddings even for closely related languages. Evaluated on two challenging datasets—Amazon Massive (multilingual digital assistant utterances) and a Song dataset (music requests with frequent code-switching)—PolyLingua achieves $99.25\%$ F1 and $98.15\%$ F1, respectively, surpassing Sonnet 3.5 while using 10× fewer parameters, making it ideal for compute- and latency-constrained environments.
\end{abstract}


\section{Introduction}
Identifying a customer’s language quickly and accurately is critical for delivering a more personal and culturally respectful experience \cite{wu-etal-2025-survey, tedeschi-etal-2021-wikineural-combined}. This need is especially acute in tools such as virtual assistants, chatbots, and customer‑service centers that interact with speakers of many languages. Detecting the language at the earliest point in the interaction enables a personalized experience by routing the user to the most appropriate content or support channel exactly when it is needed, thereby reducing friction, increasing satisfaction, and improving overall engagement and conversion rates \cite{jauhiainen2019automatic}. Moreover, early language detection allows downstream language models to be prompted in the user’s language and cultural context, making the generated responses more helpful and respectful \cite{deroy2024prompt}.

Commercial APIs and rule-based approaches \cite{nakatani2010langdetect} are widely used for language detection because of their speed and simplicity. However, they frequently struggle with short utterances or when the text contains names or entities from another language \cite{joulin2017bag}. At the same time, more precise approaches rely on large, highly resource-intensive models that are not ideal for systems with limited memory or latency. Despite its importance, language detection is frequently overlooked when developing better model architectures \cite{10.1613/jair.1.11675}. \textcolor{black}{Although large language models (LLMs) often achieve higher accuracy, they are computationally more expensive. In contrast, encoder‑only models are especially well‑suited for task‑specific problems such as sentiment analysis and named‑entity recognition \cite{zaratiana-etal-2024-gliner}, as demonstrated by recent work on ModernBERT \cite{warner-etal-2025-smarter}.}

In this paper, we frame the problem as multi-class classification because out-of-domain (OOD) languages can misleadingly resemble supported languages and be classified with high confidence. Therefore, we introduce \textsc{PolyLingua}, a lightweight yet effective multi-task model for language identification \cite{aghajanyan2021muppet, aribandi2022ext5,liu2021pretrain}. PolyLingua performs two tasks simultaneously: (1) in-domain detection to determine whether a given utterance falls under a set of supported languages, serving as a quick gating mechanism to route the request, and (2) fine-grained language classification within those supported languages to enable precise customized experiences. Both tasks are trained end-to-end with a shared encoder, allowing the model to efficiently learn common representations while supporting task-specific objectives. This shared architecture leverages the complementary information between the two tasks  to enhance generalization, while reducing the number of parameters and minimize computational overhead.

PolyLingua employs a two-level contrastive learning strategy \cite{Wu_2018_CVPR, Wang_2021_CVPR}. At the instance level, it utilizes a margin-based supervised contrastive loss for separating embeddings from confusable language pairs. At the class level, the model introduces a class-aware, margin-based separation loss that encourages each utterance to stay close to the mean embedding of its own language class while maintaining a minimum margin from those of other classes, particularly for analogous or confusable language pairs. This dual contrastive signal helps the model in forming tight, well-separated clusters in the embedding space, increasing both accuracy and generalization \cite{chen2020simple}.

Finally, we conduct thorough assessments of PolyLingua against a variety of baselines, such as commercial APIs, rule-based tools, open-source multilingual models, and LLMs \cite{anthropic2023claude} employed for language detection. \textcolor{black}{Our experiments show that PolyLingua matches the performance of LLMs while using far fewer computational resources, making it a practical choice for real‑world applications that need both accuracy and low latency.}
In summary, our contributions are:
\begin{itemize}
    \item We propose PolyLingua, a lightweight, end-to-end multi-task model that leverages a shared encoder to jointly perform in-domain detection and fine-grained language classification, avoiding the risk of confidently misclassifying OOD utterances as supported languages.
    \item We create a two-level contrastive training objective by combining instance-level supervision with a novel class-level separation loss that improves robustness to ambiguous languages.
    \item We present a comprehensive empirical evaluation of different methods using a variety of datasets. Results show that PolyLingua delivers high performance at a low computational cost.
\end{itemize}

\section{Related Works}
Recent research on multi-task learning has shown that it is capable of significantly improving sample efficiency and generalizing in low-resource and multilingual settings \cite{sanh2022multitask,aribandi2022ext5}. In parallel, contrastive learning has become a popular framework for training models with more structured and discriminative embeddings. Beyond SimCLR \cite{chen2020simple}, efforts like SupCon \cite{khosla2020supervised}and SPCL \cite{wang2022spcl} introduce supervised contrastive mechanisms that improve performance, robustness, and generalization. 

\textcolor{black}{As multilingual user bases expand, services such as virtual assistants, chatbots, and support platforms should accurately detect a user’s language to deliver personalized and culturally appropriate responses. Large models are sometimes too heavy or expensive for real‑time use, and many commercial or rule‑based solutions lack robustness}. There are several approaches in the literature for language identification. For example, FastText \cite{joulin2017bag}, langdetect \cite{nakatani2010langdetect}, and XLM-RoBERTa \cite{conneau-etal-2020-unsupervised} are popular options. However, these methods often perform poorly on short or less structured text containing code-switching languages. While prompt-based LLMs offer higher accuracy \cite{touvron2023llama, anthropic2023claude}, their computational overhead limits deployment in latency-sensitive or resource-constrained environments \cite{wan2023efficient, bai2024beyond}. In this paper, we propose PolyLingua, a model that balances speed and performance for efficient and real-world uses. 

\section{Methodology}
Let assume the training dataset consist of utterance and label tuples $\{(u_i, y_i)\}_{i=1}^N$, where $u_i$ is the input utterance, $y_i \in \mathcal{Y}$ is the language label (e.g., English, Spanish, etc.), and $d_i \in \{0, 1\}$  is a binary label indicating whether the utterance belongs to a supported in-domain language $d_i = 1$ (in-domain) or $d_i = 0$ (out-of-domain). Note that $d_i$ can be directly inferred from $y_i$, as the set of in-domain languages $\mathcal{Y}_{\text{in}} \subset \mathcal{Y}$ is known. Leveraging this, we include $d_i$ during training to enable effective multitask supervision. Below, we propose PolyLingua, a multi-task learning framework that learns more robust and separable representations to enhance language identification task. 

\subsection{Architecture}
\begin{figure}[!t]
  \centering
  \includegraphics[width=1.05\linewidth]{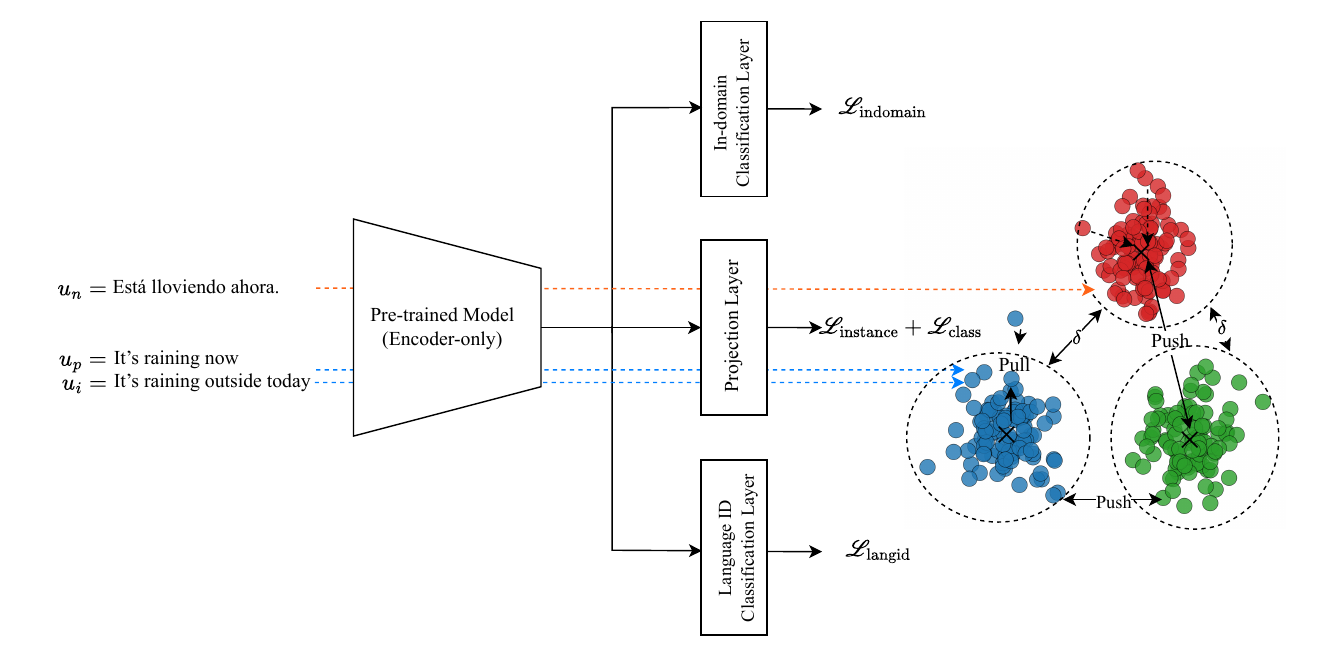}
  \caption{Given $u_i$ and a positive $u_p$ from the same class, and a negative $u_n$ from a different language, PolyLingua aligns $u_i$ with $u_p$ while pushing it away from $u_n$, using a shared encoder and class-aware margin-based contrastive losses.
}
  \label{fig:architecture}
\end{figure}

Contrastive learning is an approach that brings samples in the same class closer and pushes ones across classes further in the embedding space \cite{khosla2020supervised} (see ~\ref{eq:instance-contrastive-dot} for the objective). Leveraging this, we propose PolyLingua, a model based on two-level supervised contrastive objectives with adjustable inter-class margins, which improves performance on both in-domain detection and fine-grained language identification. The model is built based on a shared transformer encoder that converts the input utterance $u_i$ into a representation. Based on this, we incorporate three components in this encoder: (1) an in-domain detection head that performs binary classification over $d_i$; (2) a language identification head that classifies among the in-domain language classes $y_i \in \mathcal{Y}_{\text{in}}$; and (3) a projection head that converts the encoder outputs into a $L_2$-normalized embedding space (see Figure \ref{fig:architecture}). This projection is used to calculate a contrastive loss that pushes utterances from different classes away while encouraging utterances with the same language label to be near together in the embedding space.

\subsection{Training Objectives}
PolyLingua employs a two-level supervised contrastive loss at the instance level. The loss for the projected embeddings $\{z_i\}$ pushes apart different-label tuples with class-awareness that promoting proximity between same-label pairs.
\begin{equation}\label{eq:instance-contrastive-dot}
\begin{split}
\mathcal{L}_{\text{instance}} = \sum_{i \in \mathcal{I}} \frac{1}{|P(i)|} \sum_{p \in P(i)} -\log \Bigg(
\frac{\exp(z_i^\top z_p / \tau)}{\sum_{a \in A(i)} \exp(z_i^\top z_a / \tau)}
\Bigg),
\end{split}
\end{equation}
where $\tau > 0$ is the temperature, $\mathcal{I}$ represents the set of all anchor instances, and $P(i)$ represents the set of indices corresponding to positive utterances for instance $i$, encouraging proximity between pairs with the same label. $A(i)$ stands for the set of all candidate samples used in the batch except $i$. 

In addition to instance-level contrastive learning, we enforce separation at the class level by encouraging each embedding $z_i$ to be close to the mean embedding of its class and far from those of other classes. Let $c_y$ denote the centroid (mean embedding) of class $y$, computed as:
\begin{equation}
c_y = \frac{1}{|A_y|} \sum_{i \in A_y} z_i,
\end{equation}
where $A_y = \{i \in \mathcal{I} : y_i = y\}$ is the set of samples belonging to class $y$ in the batch. We then define a class-level contrastive loss:

\begin{equation}\label{eq:class-contrastive-dot}
\begin{split}
\mathcal{L}_{\text{class}} = \sum_{i \in \mathcal{I}} -\log \Bigg(
\frac{\exp(z_i^\top c_{y_i} / \tau)}{\sum_{y \in \mathcal{Y}_{\text{in}}} \exp(z_i^\top c_y / \tau - \delta_{y_i, y})}
\Bigg),
\end{split}
\end{equation}
where $\delta_{y_i, y} \geq 0$ is an adaptive margin penalty between classes $y_i$ and $y$.

\textbf{Adaptive margins for linguistically similar classes}: A key contribution of our approach is the introduction of class-pair-specific adaptive margins that reflect linguistic similarity between languages. Linguistically similar languages (e.g., Portuguese and Spanish, which share substantial vocabulary and grammatical structures, or French and Spanish with considerable lexical overlap) are more prone to confusion. To enforce stronger separation between such confusing pairs, we assign larger margins:

\begin{equation}
\delta_{y_i, y} = \begin{cases}
0 & \text{if } y_i = y \\
\delta_{\text{high}} & \text{if } y_i \neq y \text{ and } (y_i, y) \in \mathcal{P}_{\text{confusing}} \\
\delta_{\text{low}} & \text{otherwise}
\end{cases}
\end{equation}
where $\delta_{\text{high}} > \delta_{\text{low}} \geq 0$, $\mathcal{P}_{\text{confusing}}$ is the set of confusing language pairs identified based on linguistic family relationships and empirical confusion patterns. The margin penalty $\delta_{y_i, y}$ reduces the similarity score for negative class centroids in the denominator of Equation~\ref{eq:class-contrastive-dot}, making $\exp(z_i^\top c_y / \tau - \delta_{y_i, y})$ smaller for negative classes. This effectively decreases their contribution to the denominator, making it harder for the model to assign high probabilities to incorrect classes, particularly those linguistically similar to the true class. During training, this mechanism pushes each embedding closer to its class centroid while enforcing larger separations from confusing classes, resulting in well-separated and discriminative language clusters in the embedding space.

\textbf{Combined objective:} The final training objective combines both contrastive losses with the standard cross-entropy losses for in-domain detection and language identification:

\begin{equation}\label{eq:total-loss}
\mathcal{L} = \lambda_1 \mathcal{L}_{\text{indomain}} + \lambda_2\mathcal{L}_{\text{langid}} + \lambda_3 (\mathcal{L}_{\text{instance}} + \mathcal{L}_{\text{class}}),
\end{equation}

where $\mathcal{L}_{\text{indomain}}$ and $\mathcal{L}_{\text{langid}}$ are cross-entropy losses for the domain detection and language identification heads respectively, and $\lambda_1, \lambda_2$ are weighting hyperparameters that balance the contribution of each loss component.


\begin{figure*}[ht]
    \centering
    \includegraphics[width=\textwidth, trim=70pt 70pt 70pt 70pt, clip]{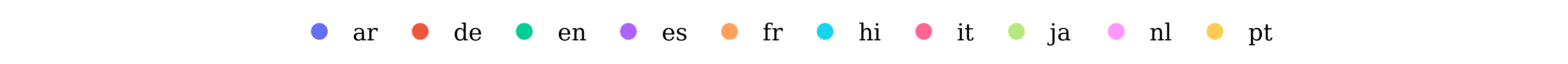}
    \begin{subfigure}[t]{0.32\textwidth}
        \centering
        \includegraphics[width=\linewidth]{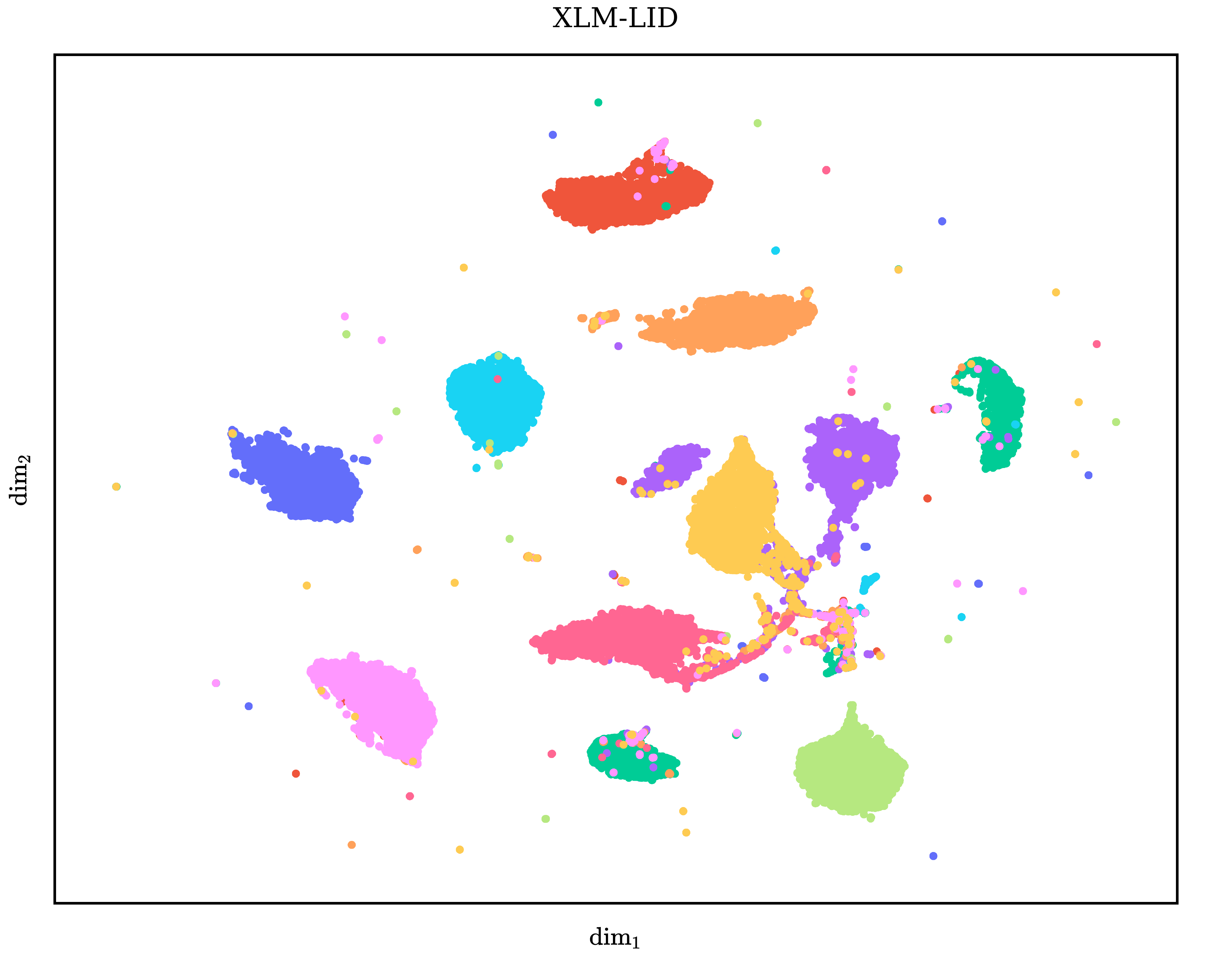}
        \caption{XLM-LID on Amazon Massive}
        \label{fig:umap-amazon-xlm}
    \end{subfigure}
    \hfill
    \begin{subfigure}[t]{0.32\textwidth}
        \centering
        \includegraphics[width=\linewidth]{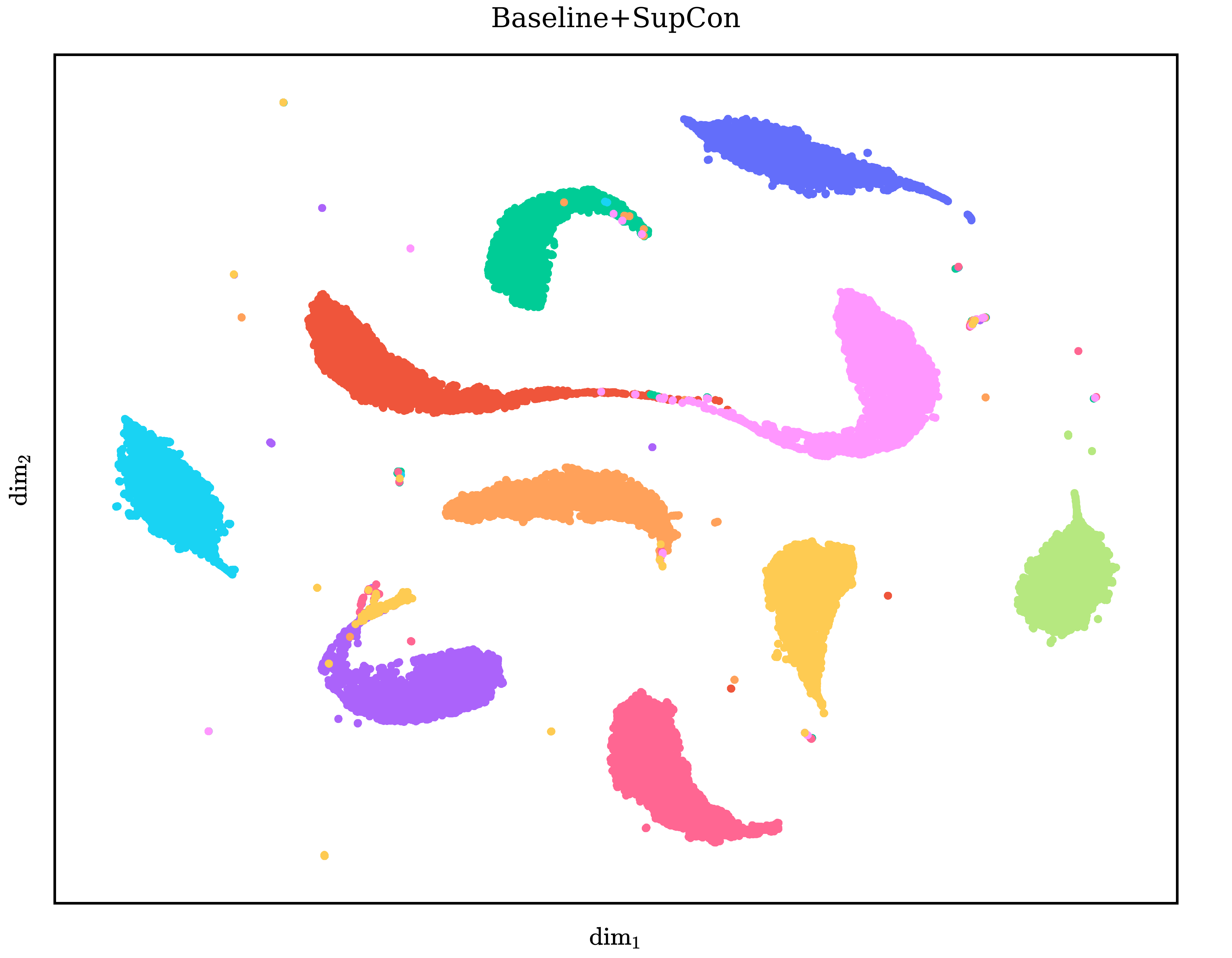}
        \caption{Baseline+SupCon on Amazon Massive}
        \label{fig:umap-amazon-baseline}
    \end{subfigure}
    \hfill
    \begin{subfigure}[t]{0.32\textwidth}
        \centering
        \includegraphics[width=\linewidth]{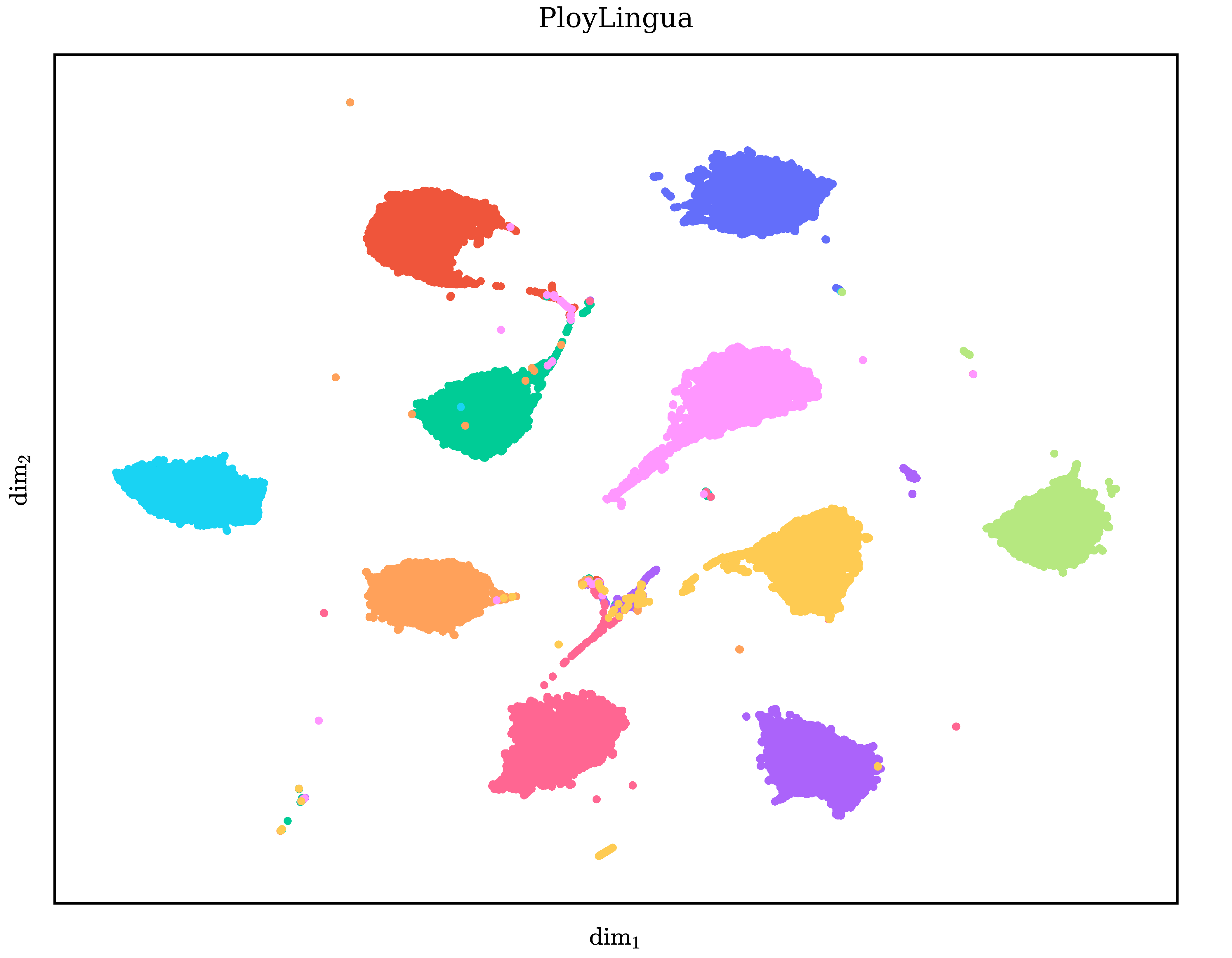}
        \caption{PolyLingua on Amazon Massive}
        \label{fig:umap-amazon-poly}
    \end{subfigure}
    \begin{subfigure}[t]{0.32\textwidth}
        \centering
        \includegraphics[width=\linewidth]{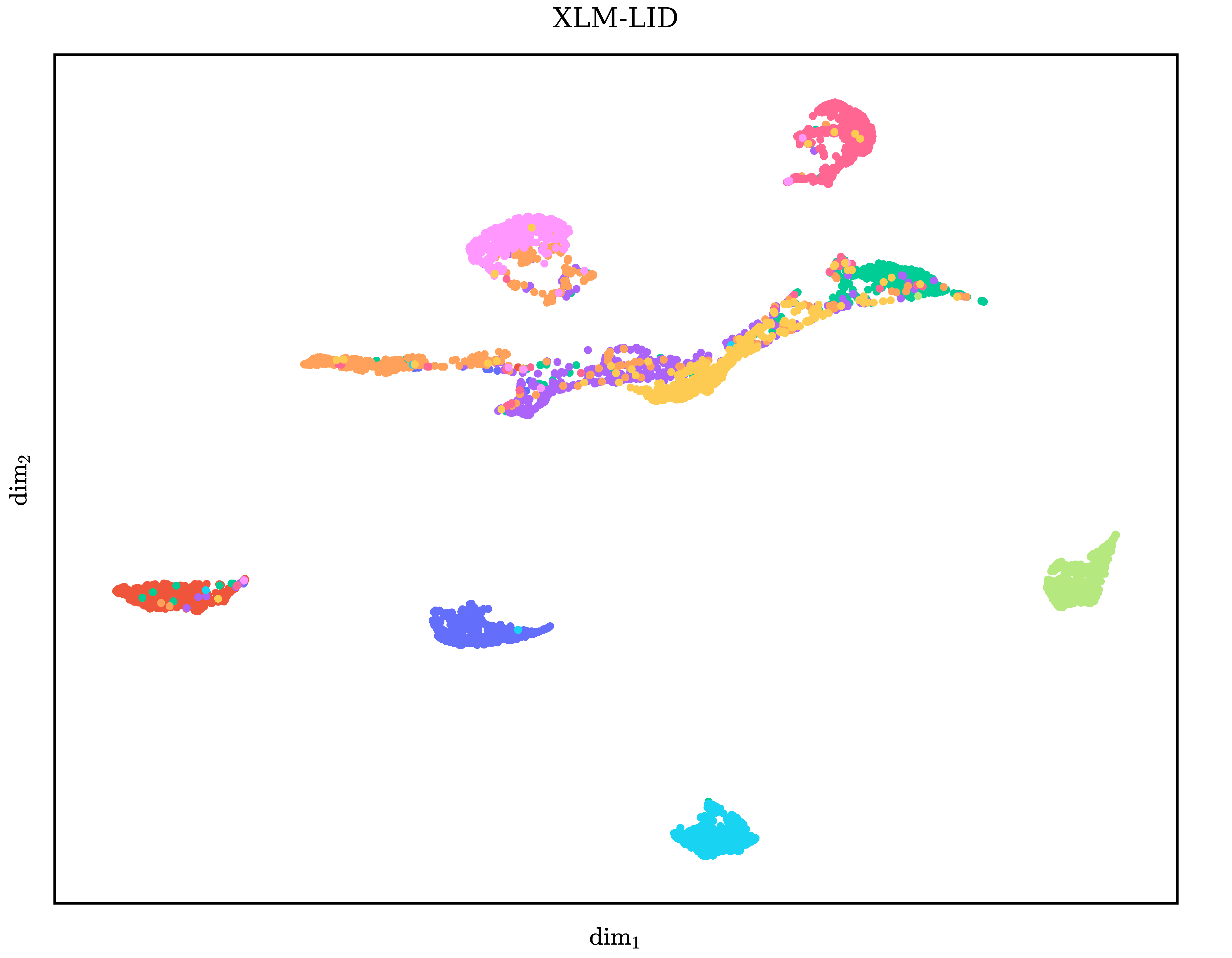}
        \caption{XLM-LID on Song Dataset}
        \label{fig:umap-song-xlm}
    \end{subfigure}
    \hfill
    \begin{subfigure}[t]{0.32\textwidth}
        \centering
        \includegraphics[width=\linewidth]{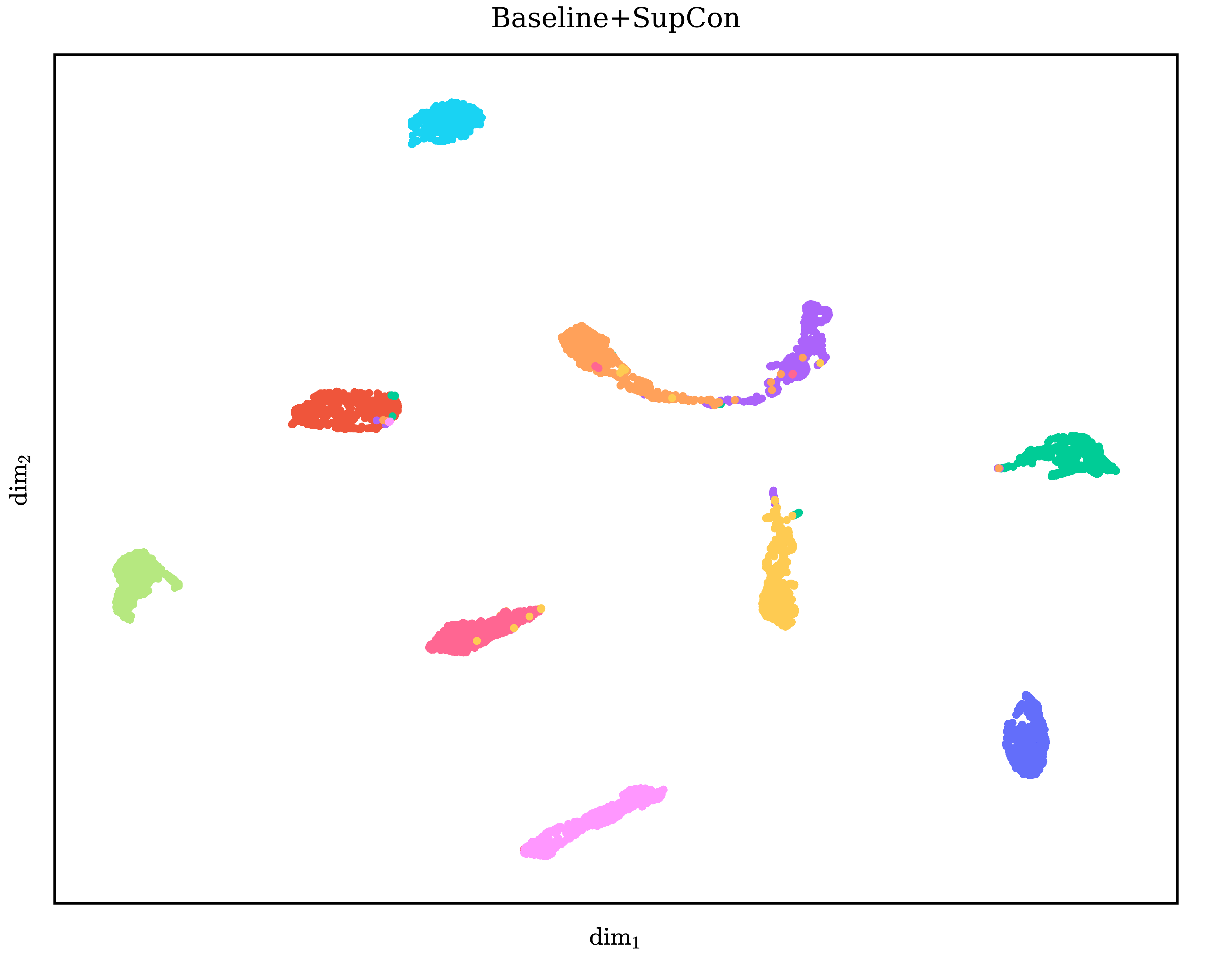}
        \caption{Baseline+SupCon on Song Dataset}
        \label{fig:umap-song-baseline}
    \end{subfigure}
    \hfill
    \begin{subfigure}[t]{0.32\textwidth}
        \centering
        \includegraphics[width=\linewidth]{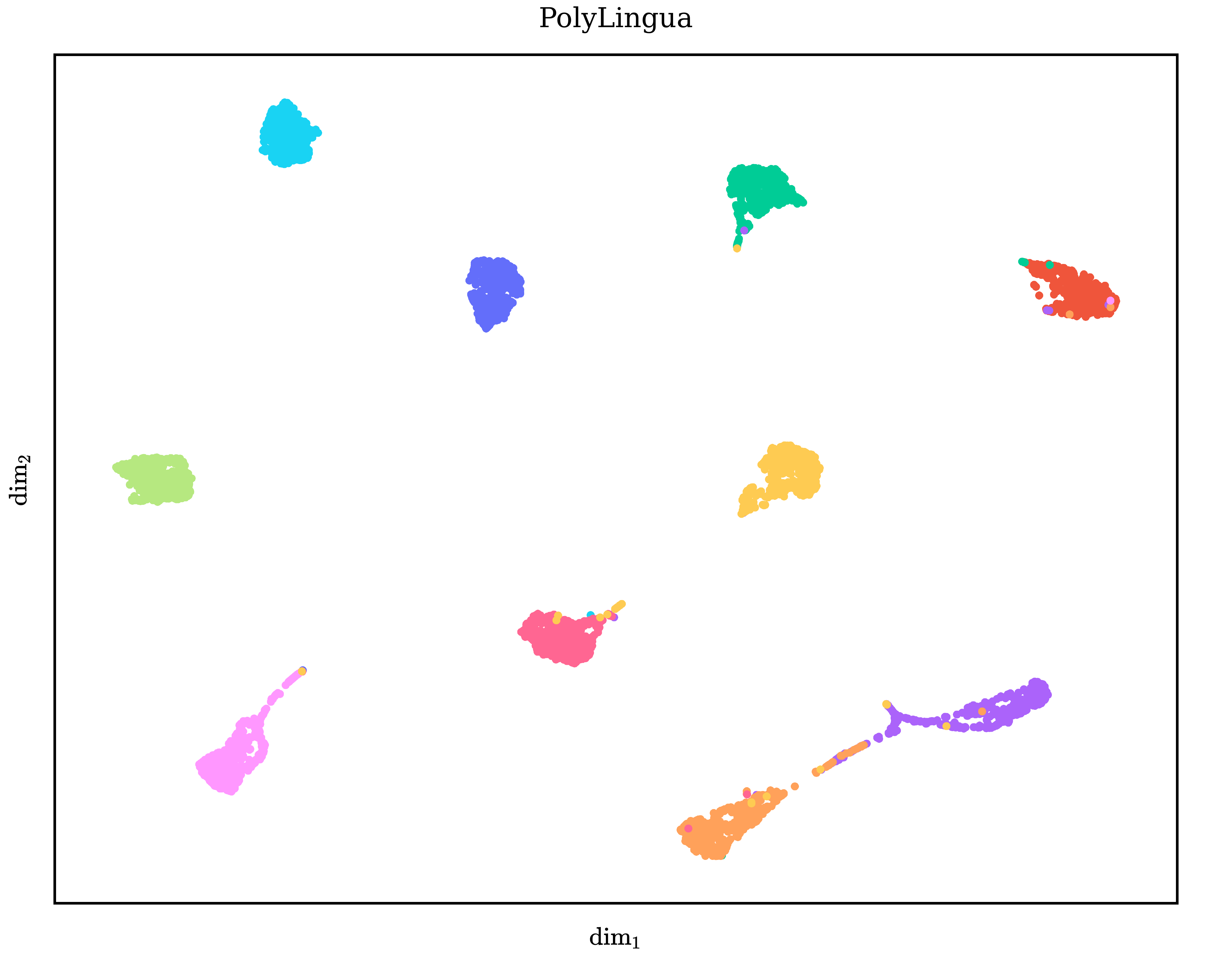}
        \caption{PolyLingua on Song Dataset}
        \label{fig:umap-song-poly}
    \end{subfigure}

    \caption{
    UMAP projections of utterance embeddings from Amazon Massive (top row) and the Song dataset (bottom row). Each point represents an utterance and is colored by its true language label. On Amazon Massive, PolyLingua forms more compact and well-separated clusters than XLM-LID and Baseline+SupCon, especially for similar languages such as French, Portuguese, and Spanish, due to its margin-based inter-class separation. On the Song dataset, which includes noisy utterances with diverse and multilingual artist and song entities, PolyLingua again shows clearer clustering, demonstrating robustness to intra-class variation and entity-induced noise.
    }
    \label{fig:umap_combined}
\end{figure*}



\text{Positive pairs}: To generate positive pairs for contrastive learning, we utilized data augmentation techniques that maintain the original semantics. An original utterance and its augmented variant produced using the following techniques make up each positive pair:
\begin{itemize}
    \item \textbf{Random Deletion}: Randomly removing tokens (with probability of $p = 0.15$).
    \item \textbf{Adding Noise}: Injects character-level perturbations typing errors and noisy user inputs or ASR transcriptions.
    \item \textbf{Dynamic Entity Replacement}: Using a multilingual named entity recognition model~\cite{tedeschi-etal-2021-wikineural-combined} to locate entities in the utterance with high confidence. Then in each epoch, we replaced another randomly chosen entity of the same type. This increases cross-lingual diversity while preserving semantic consistency.
\end{itemize}
Further information on data augmentation and hyperparameter tuning is provided in Appendix A.

\begin{table*}[ht]
\centering
\sisetup{detect-weight=true}  
\caption{
Comparison of models performance on the Amazon Massive and Song datasets.
Metrics reported: In-domain Accuracy (InAcc), Precision (Prec), Recall, Top-1 Accuracy (Top-1), Top-5 Accuracy (Top-5), and F1 Score, all in percentage (\%).
}
\label{tab:results-accuracy}
\small
\renewcommand{\arraystretch}{1} 
\setlength{\tabcolsep}{6.5 pt} 
\begin{tabular}{
  l
  S[table-format=2.1] S[table-format=2.1] S[table-format=2.1] S[table-format=2.1] S[table-format=2.1] S[table-format=2.1]
  S[table-format=2.1] S[table-format=2.1] S[table-format=2.1] S[table-format=2.1] S[table-format=2.1] S[table-format=2.1]
}
\toprule
\multirow{2}{*}{\textbf{Model}} 
& \multicolumn{6}{c}{\textbf{Amazon Massive}} 
& \multicolumn{6}{c}{\textbf{Song Dataset}} \\
\cmidrule(lr){2-7} \cmidrule(lr){8-13}
& \rotatebox{60}{InAcc} & \rotatebox{60}{Prec} & \rotatebox{60}{Recall} & \rotatebox{60}{Top-1} & \rotatebox{60}{Top-5} & \rotatebox{60}{F1}
& \rotatebox{60}{InAcc} & \rotatebox{60}{Prec} & \rotatebox{60}{Recall} & \rotatebox{60}{Top-1} & \rotatebox{60}{Top-5} & \rotatebox{60}{F1} \\
\midrule
Langdetect           & 95.08 & 79.17 & 79.29 & 79.29 & {--} & 78.60 & 90.25 & 71.33 & 40.13 & 40.13 & {--} & 41.18 \\
FastText             & 97.74 & 94.56 & 92.95 & 92.95 & {--} & 93.20 & 95.44 & 89.60 & 79.44 & 79.44 & {--} & 81.53  \\
XLM-LID              & 82.59 & 97.43 & 96.02 & 96.02 & 98.64 & 96.65 & 91.45 & 87.02 & 82.18 & 82.18 & 90.06 & 81.73 \\
Baseline             & \bfseries 99.58 & 99.20 & 99.20 & 99.20 & \bfseries 99.52 & 99.20 & 96.24 & 95.41 & 91.83 & 91.83 & 98.84 & 93.47 \\
Baseline+SupCon      & 99.27 & 99.24 & 99.23 & 99.23 & 99.47 & 99.23 & 97.82 & 95.79 & 93.15 & 93.15 & 96.60 & 94.31 \\
\rowcolor{gray!20} Sonnet 3.5.          & 99.67 & 97.90 & 97.77 & 97.77 & {--} & 97.79 & 98.95 & 96.07 & 95.03 & 95.03 & {--} & 95.52 \\
\textbf{PolyLingua}  & 99.51 & \bfseries 99.25 & \bfseries 99.25 & \bfseries 99.25 & 99.49 & \bfseries 99.25
                     & \bfseries 99.71 & \bfseries 98.29 & \bfseries 98.06 & \bfseries 98.06 & \bfseries 99.73 & \bfseries 98.15 \\
\bottomrule
\end{tabular}
\end{table*}

\section{Experiments}
\subsection{Model Setup}
We train our model in a multitask setup consisting of two classification objectives. We use a single backbone with two classification heads for each objective. As a result, sharing a common base model for both tasks simplifies deployment and lowers latency during inference. Additionally, we used a projection head with a margin-based, class-aware loss to help the model group examples from the same language together and push apart different ones. This made it easier to tell apart similar languages like Portuguese, Spanish, and French. For experiments, we employed the multilingual MiniLM encoder \cite{gu2023minillm} due to its efficiency and robust performance across a wide range of languages. The encoder processes each utterance and produces the embeddings. Lastly, the model is trained end-to-end using the joint objective presented in \ref{eq:total-loss}.

\subsection{Datasets}
\subsubsection{Amazon Massive}
We use the Amazon Massive dataset, which covers 52 languages on natural language understanding (NLU) tasks. In our experiments, we only focus on the utterances and their language labels to build a dataset for both in-domain detection and language classification. For this, we pick 10 in-domain target languages for classification: English, Spanish, French, Arabic, Hindi, Dutch, German, Italian, Portuguese, and Japanese, which are the key languages for our international businesses. Therefore, utterances from the remaining languages are considered out-of-domain (OOD) and are used for the in-domain detection task. To preserve class balance as the original dataset, we subsample the out-of-domain utterances such that the out-domain examples constitute 40\% of the total training set size. Additionally, we follow the same approach to create the test set from the Amazon Massive test split. 
More information about the dataset has been provided in Appendix C.

\subsubsection{Synthetic Dataset}
To test our model's ability to handle real and less structured inputs, we also constructed a difficult synthetic dataset containing a lot of entities that mimics real-world conversations, such as music requests with artist and song names. For each in-domain language, we created 50 different diverse templates with placeholders like \textit{[SONG\_NAME]} and \textit{[ARTIST\_NAME]}. Eventually, we filled these placeholders with actual names from the Million Song Dataset \cite{Bertin-Mahieux2011}. As a result, the resulting dataset covers a broad range of entities; spanning several languages, genres, and scripts (such as non-standard capitalization and accented characters). This made the utterances complex and often ambiguous. Moreover, these templates reflect informal grammar, code-switching, and incomplete structures common in real-world utterances. We eventually created 10,000 of these utterances per language
that many language identification techniques currently in use find challenging to tag. See Appendix C for more information. 

\subsection{Base Models}
We tested our proposed model against a variety of language detection methods and baselines: (1) \textbf{Baseline}: A multi-task model trained with just cross-entropy (CE) loss for ablation analysis, using the same architecture as our proposed method; (2) \textbf{Baseline+SupCon}: A multi-task model trained with the same architecture, using cross-entropy loss for both tasks along with a supervised contrastive loss ~\cite{conneau2020unsupervised}, allowing us to evaluate the contribution of the proposed loss functions; (3) \textbf{FastText (LID.176.bin)}: An efficient, open-source language identifier trained on Wikipedia text \cite{joulin2017bag}; (4) \textbf{Langdetect}: A lightweight, rule-based language detection tool \cite{nakatani2010langdetect}; (5) \textbf{XLM-LID}: A multilingual transformer model fine-tuned for language identification, available at \textit{papluca/xlm-roberta-base-language-detection}; and (6) \textbf{Sonnet 3.5 (via Amazon Bedrock)}: A powerful large language model prompted with a carefully crafted instruction for language detection. See Appendix B for additional details about the prompt.

These baselines include a wide variety of simple rule-based tools, open-source models, commercial APIs, and  large language models, enabling us to compare our approach from different angles.

\subsection{Results}
The results in Table ~\ref{tab:results-accuracy} compare multiple language identification models evaluated on the two datasets. Conventional tools such as Langdetect and FastText show reasonable performance, with FastText outperforming Langdetect notably in both precision and recall on the Amazon dataset. XLM-LID shows high recall but has significantly lower in-domain accuracy (82.59\%). Our two ablation models—Baseline and Baseline+SupCon—show consistent improvement over open-source and commercial tools, with top-5 accuracy exceeding 99\% in both variants, suggesting that supervised contrastive learning contributes positively to model robustness. For additional comparative examples of PolyLingua versus Sonnet 3.5, see Table ~\ref{tab:code_mixed_examples} in the Appendix C.

Our proposed model, PolyLingua, achieves the best overall performance across both datasets. On Amazon Massive, it has a top-1 accuracy of 99.25\% and  the highest F1 score of 99.25\%. More importantly, the improvements are even more prominent on the more challenging Song Dataset. Additionally, Polylingua surpasses all baselines with an F1 score of 98.15\%, indicating strong generalization across unseen and noisier utterances. Compared to Sonnet 3.5, for which we constructed a strong tailored prompt (see Appendix C), PolyLingua delivers higher accuracy and recall, particularly in top-5 metrics (99.73\% vs. 96.60\%). Overall, these findings show that PolyLingua performs better than LLM, commercial APIs, and supervised baseline. This highlights the efficacy of our end-to-end training.  Also, we report on average latency per single inference in Table ~\ref{tab:latency_only}, with full benchmarking details provided in Appendix A.

\begin{figure*}[ht]
    \centering
    \begin{subfigure}[t]{0.32\textwidth}
        \centering
        \includegraphics[width=\linewidth]{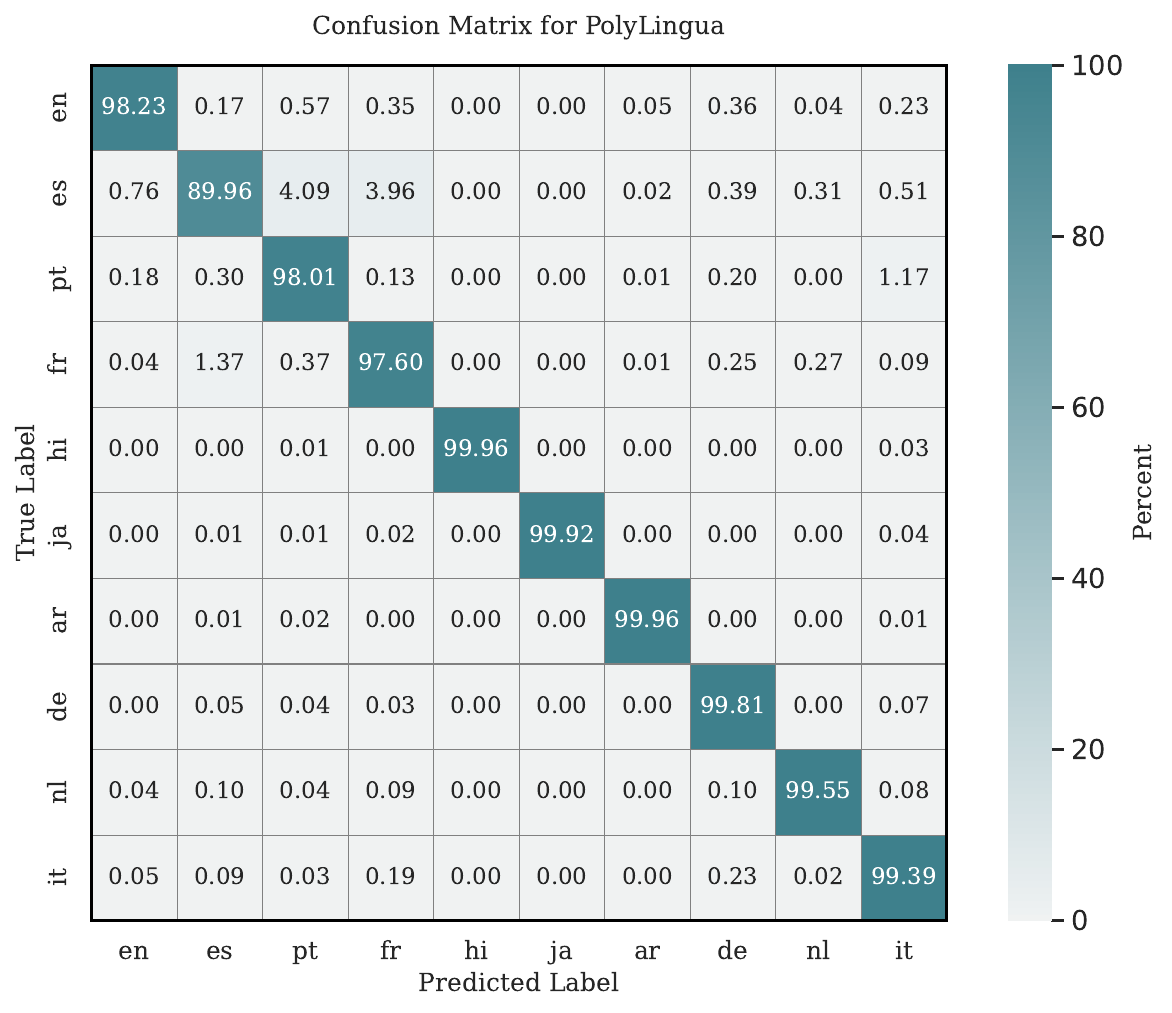}
        \caption{PolyLingual Confusion Matrix}
        \label{fig:confusion}
    \end{subfigure}%
    \hfill
    \begin{subfigure}[t]{0.32\textwidth}
        \centering
        \includegraphics[width=\linewidth]{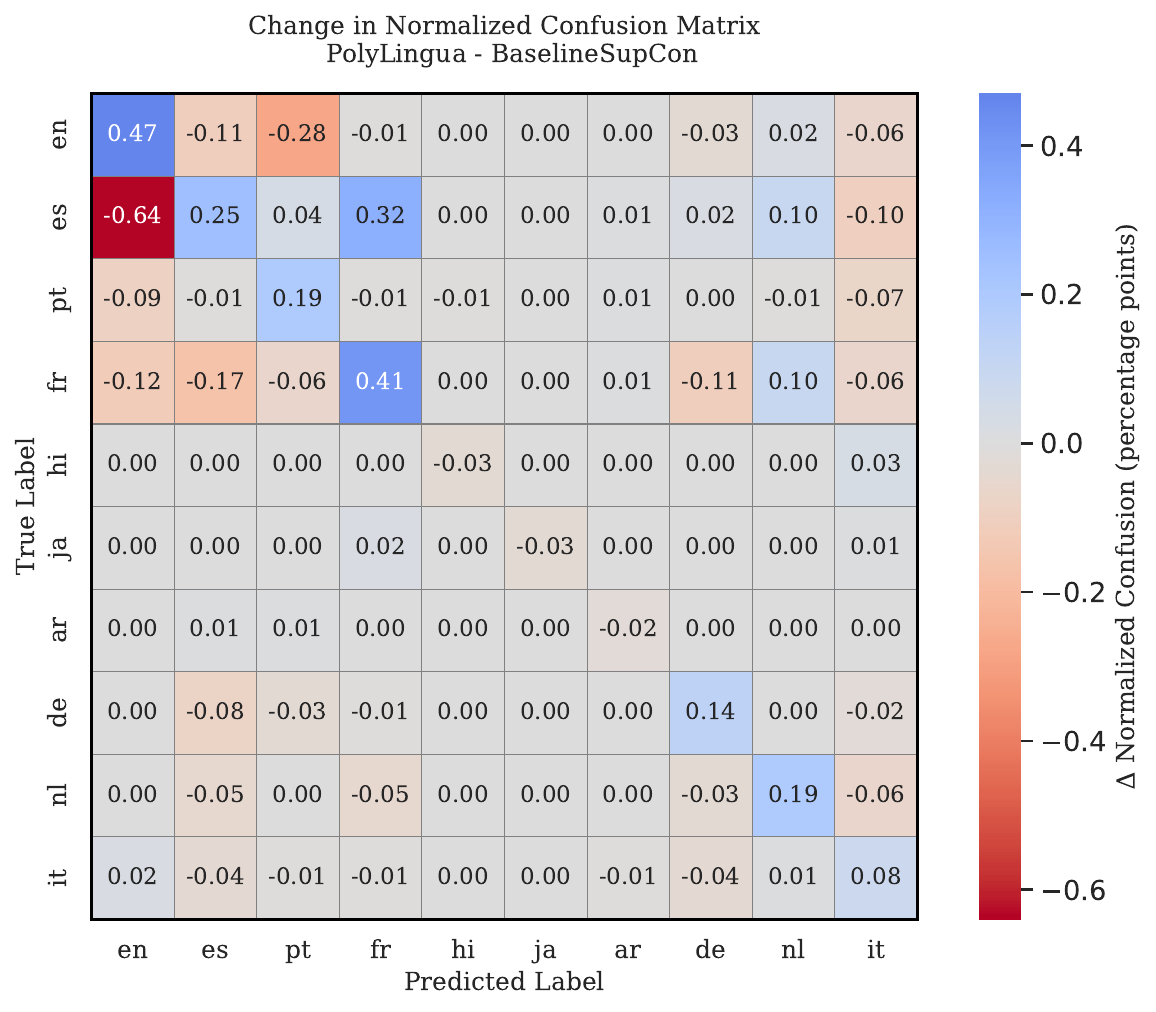}    
        \caption{Change in Confusion: PolyLingua -- BaselineSupCon}
        \label{fig:baselineconfusion}
    \end{subfigure}%
    \hfill
    \begin{subfigure}[t]{0.32\textwidth}
        \centering
        \includegraphics[width=\linewidth]{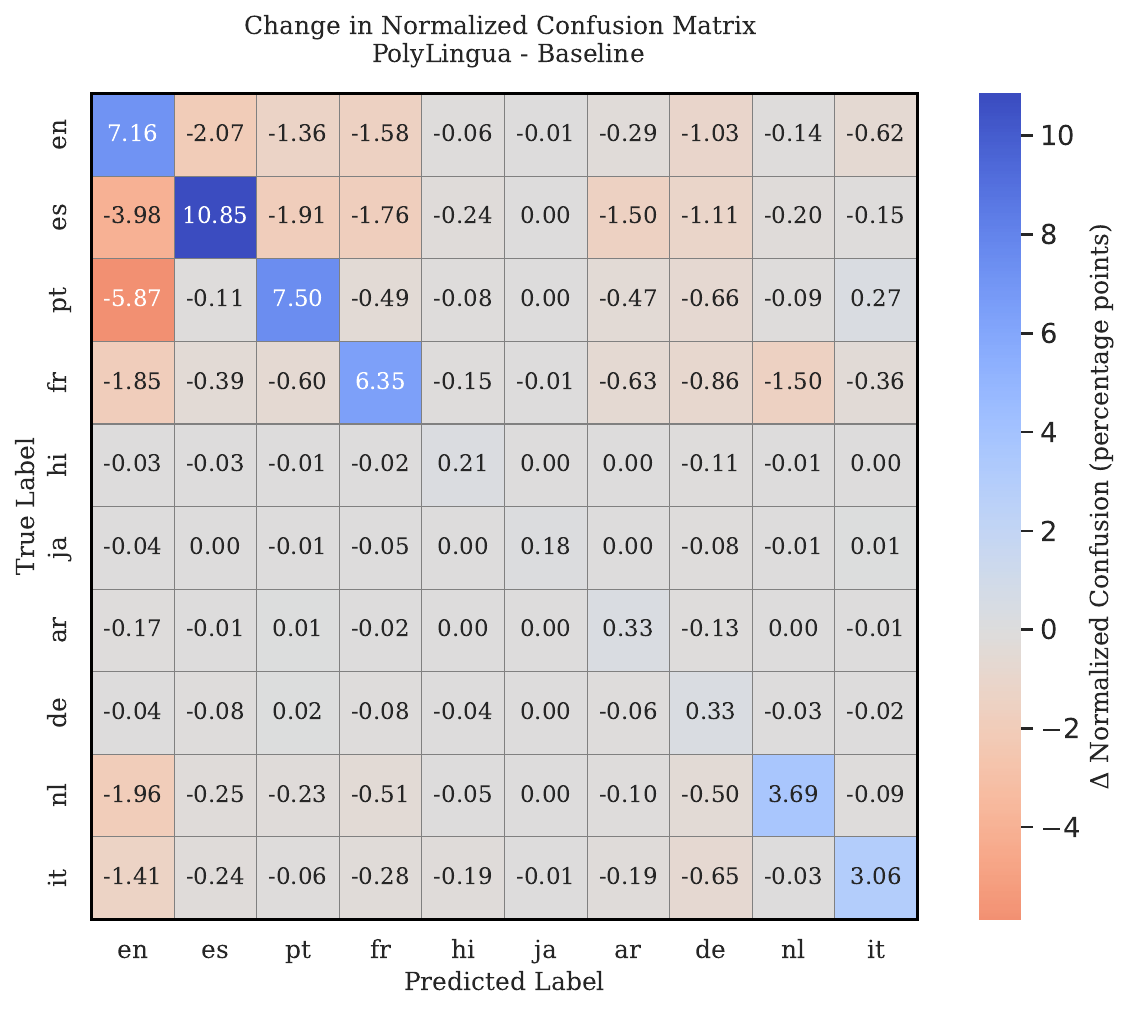}
        \caption{Change in Confusion: PolyLingua -- Baseline}
        \label{fig:Confusion-Matrix}
    \end{subfigure}
    
    \caption{\textbf{Left}: Confusion matrix for the proposed PolyLingua model on 10 in-domain languages, demonstrating low misclassification.
    \textbf{Center}: Difference in the normalized confusion matrices between PolyLingua and the Baseline+SupCon model.
    \textbf{Right}: Difference between PolyLingua and the standard Baseline. Blue cells on the diagonal indicate improvements in true positive rates by PolyLingua, while red cells off the diagonal represent reductions in misclassification and confusion. Arrow pointer indicates the improvement in the performance in the Difference confusion matrix \protect\includegraphics[width=1.3em]{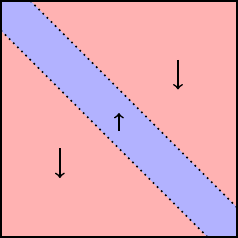}.}
    \label{fig:confusions-matrix}
\end{figure*}

\begin{table}[t]
\centering
\caption{Average latency per inference call for different models.}
\label{tab:latency_only}
\small
\setlength{\tabcolsep}{3pt} 
\begin{tabular}{l S[table-format=4.2, table-align-text-post=false, table-number-alignment=right]}
\\
\toprule
\textbf{Model} & {\textbf{Latency (ms)}} \\
\midrule
Langdetect      & 8.53 \\
FastText        & 0.01 \\
XLM-LID         & 7.29 \\
Sonnet 3.5      & 1373.31 \\
\textbf{PolyLingua}     & \bfseries 7.19 \\
\bottomrule
\end{tabular}
\end{table}

\begin{figure*}[ht]
    \centering
    \begin{subfigure}[t]{0.32\textwidth}
        \centering
        \includegraphics[width=\linewidth]{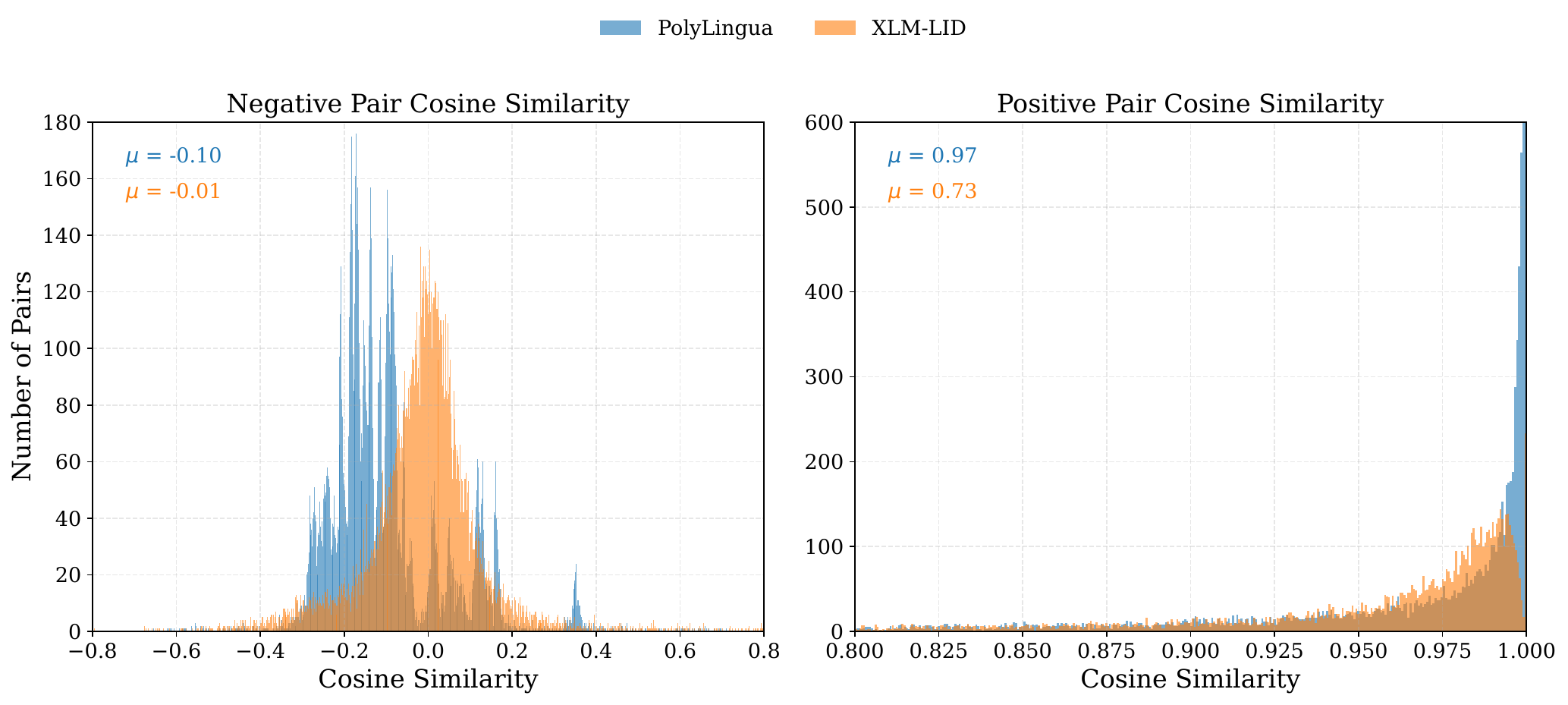}
        \caption{Cosine similarity for positive and negative pairs: \\ PolyLingua vs XLM-LID.}
        \label{fig:ploylingua_xlm_cosine}
    \end{subfigure}
    \hfill
    \begin{subfigure}[t]{0.32\textwidth}
        \centering
        \includegraphics[width=\linewidth]{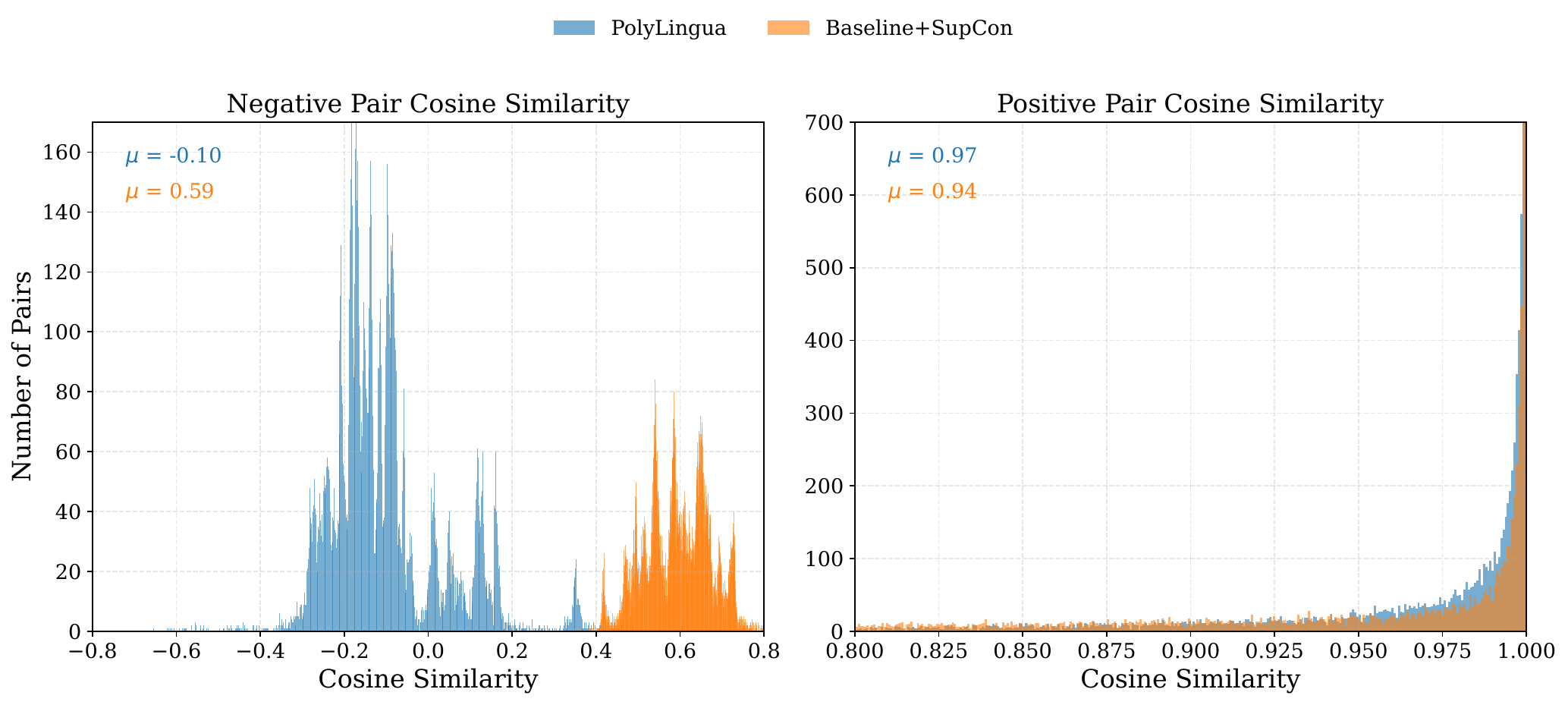}
        \caption{Cosine similarity for positive and negative pairs: \\PolyLingua vs Baseline+SupCon.}
        \label{fig:ploylingua_baseline_cosine}
    \end{subfigure}
    \hfill
    \begin{subfigure}[t]{0.32\textwidth}
        \centering
        \includegraphics[width=\linewidth]{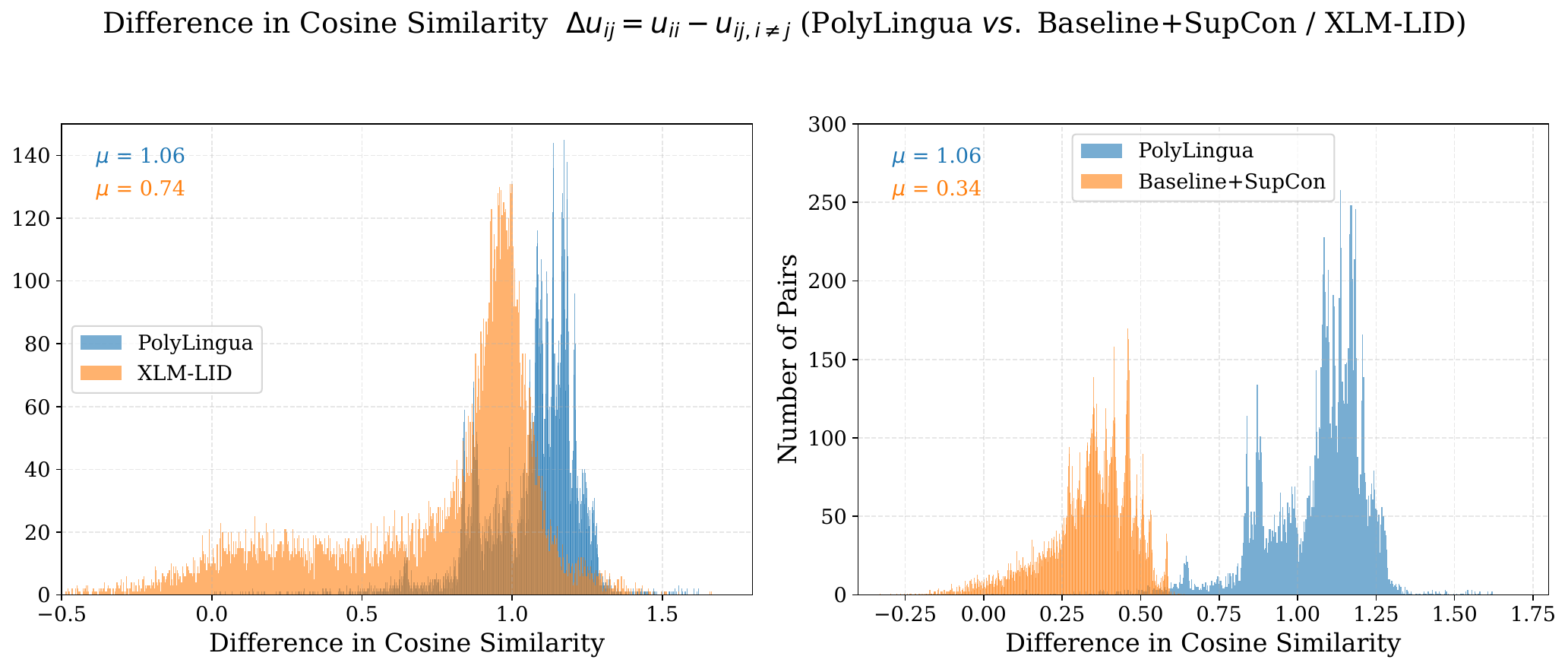}
        \caption{Difference in cosine similarity for positive vs. negative pairs: PolyLingua vs. XLM-LID/Baseline+SupCon}
        \label{fig:difference_cosine}
    \end{subfigure}
    \caption{Comparison of cosine similarity distributions between PolyLingua and baseline models on the Amazon Massive dataset. (a) and (b) show positive and negative pair distributions for PolyLingua against XLM-LID and Baseline+SupCon, respectively. (c) shows difference histograms, indicating better class separation in PolyLingua.}
    \label{fig:consine-distribution}
\end{figure*}

In Figure~\ref{fig:confusions-matrix}, left plot displays the confusion matrix for PolyLingua, indicating high true positive rates and low confusion. On the other side, the center and right plots visualize the differential impact of PolyLingua over the Baseline+SupCon and Baseline models, respectively. Blue on diagonal cells show gains in true positives, while red off-diagonal cells show reduced misclassification.
These results demonstrate that our two-level margin-based contrastive loss significantly enhances language separation and reduces confusion, especially among closely related languages such as Spanish, Portuguese, French and English. Arrows in the caption highlight improvements.

\begin{table}[ht]
\centering
\footnotesize
\setlength{\tabcolsep}{4pt}
\caption{Performance comparison on Amazon Massive dataset.}
\label{tab:amazon_Massive_results_embeddings_top1&5}
\begin{tabular}{
  l
  S[table-format=2.2] S[table-format=2.2]
  S[table-format=1.3] S[table-format=1.3] S[table-format=2.2]
}
\toprule
Method 
& {T-1↑} & {T-5↑} 
& {Inter↑} & {Intra↓} & {Ratio↑} \\
\midrule
Baseline+SupCon   & 93.98 & 95.02 & 0.042 & 0.042 & 9.72 \\
XLM-LID           & 84.64 & 88.91 & 0.242 & 0.978 & 4.04 \\
PolyLingua        & \textbf{97.69} & \textbf{98.43} & \textbf{1.096} & \textbf{0.037} & \textbf{29.79} \\
\bottomrule
\end{tabular}
\end{table}

Figure~\ref{fig:umap_combined} presents 2D UMAP projections of model embeddings on the Amazon Massive and Song datasets for visualization. In both cases, each maker is an utterance colored by its language. We can see how well each model groups same-language utterances and separates different languages. Our model, PolyLingua, gives tighter and more separated clusters compared to XLM-LID and Baseline+SupCon. Indeed, it clearly separates confusing language pairs like French, Portuguese, and Spanish. On the Song dataset, which includes artist and song names with added noise, PolyLingua still forms clean language groups. This shows it can focus on language signals even when the input is noisy.

To further assess how well different approaches distinguish between similar and dissimilar utterances, we quantify the results in Figure ~\ref{fig:consine-distribution}. The cosine similarity for both positive and negative pairings can be seen in the top histograms. Positive samples for PolyLingua are located closer together, while negative samples are broadly spaced compared to both Baseline+SupCon and XLM-LID. More importantly, the difference between the positive and negative pair similarities is displayed in the bottom plots. It is worth mentioning that better separation is indicated by a larger difference. Using this, the biggest difference is seen by PolyLingua, which demonstrates superior class boundary learning.

These findings demonstrate that PolyLingua creates clusters that are more robust and distinct than the baselines. Also, different approaches of assessing model performance are shown in Table ~\ref{tab:amazon_Massive_results_embeddings_top1&5}, which uses a k-NN classifier on the learned embeddings to measure classification accuracy.  This approach evaluates the capability of presented models to effectively capture representations without the need for a separate classifier head. With higher Top-1 and Top-5 accuracy, larger inter-class distance, and lower intra-class variation, PolyLingua performs better than Baseline and XLM-LID on all measures. Its high inter-to-intra ratio provides more evidence of its capacity to create small, distinct language clusters. 

\section{Conclusion}
We introduced PolyLingua, a lightweight multi-task model for language identification that performs in-domain detection and fine-grained classification simultaneously. PolyLingua achieves outstanding accuracy and computational efficiency by utilizing a shared encoder and a two-level contrastive learning strategy. As part of this, we propose a novel two-level margin-based contrastive loss function that brings embeddings of the same class closer and explicitly enforces separation between different classes by introducing adjustable inter-class margins at the class level. By doing this, it promotes compact clusters and improves the model’s performance to distinguish between closely related languages. 

\bibliography{aaai2026}

\appendix
\section*{Appendix A: Implementation Details}
\subsection*{Parameters}
We fine-tune a lightweight multilingual transformer, MiniLM-L12-H384, in a multi-task learning setup that performs binary in-domain detection as well as multi-class language identification. Notably, the suggested technique is model-agnostic and can be used with any other pretrained encoder. Please see Table ~\ref{tab:hyperparams} for more details. Hyperparameters, including $\lambda_1, \lambda_2, \cdots, \delta_{\text{high}}$ and others, were optimized using random search.
\begin{table*}[t]
\caption{Implementation Details}
\centering
\small
\begin{tabular}{ll}
\toprule
\textbf{Parameters} & \textbf{Values} \\
\midrule
Pretrained Model         & \texttt{Multilingual-MiniLM-L12-H384} \\
Max Input Length         & 256 tokens \\
Batch Size (per GPU)     & 150 \\
Epochs                   & 10 \\
Optimizer                & AdamW \\
Learning Rate            & 2e-5 \\
Scheduler                & CosineAnnealingLR ($T_{max} = 5$) \\
Temperature ($\tau$)     & 0.07 \\
Class Contrastive Loss Weight    & 20.0 \\
Inter-class Margin Loss Weight   & 20.0 \\
Contrastive Margin ($\delta_{\text{high}})$        & 0.4 (for pairs of \texttt{es, pt, fr}) \\
Contrastive Margin ($\delta_{\text{low}})$        & 0.0 (for other pairs)\\
Classification Loss Weight ($\lambda_1, \lambda_2$) & 1.0 (each for in-domain and language ID) \\
Contrastive Loss Weight $\lambda_3$ & 0.1  \\
Augmentations            & Random deletion, typo noise, entity replacement \\
\bottomrule
\end{tabular}
\label{tab:hyperparams}
\end{table*}

\begin{table*}[ht]
\centering
\caption{\textcolor{black}{sample code-mixed utterances from the Song Dataset where PolyLingua correctly identifies the primary language while Sonnet 3.5 misclassifies them. Utterances contain commands and song/artist names (highlighted in orange) mixing multiple languages.}}
\label{tab:code_mixed_examples}
\begin{tabular}{%
    >{\raggedright\arraybackslash}p{0.55\linewidth} 
    >{\centering\arraybackslash}p{0.1\linewidth} 
    >{\centering\arraybackslash}p{0.1\linewidth} 
    >{\centering\arraybackslash}p{0.14\linewidth}
}
\toprule
\textbf{Utterance} & \textbf{Label} & \textbf{PolyLingua} & \textbf{Sonnet 3.5} \\
\midrule
\small reproduz \textcolor{orange}{Looking Thru Bloodshot Eyes do The Casualties} & pt & pt & en \\
\small toca por favor \textcolor{orange}{G2 K1 do Tara Putra} & pt & pt & en \\
\small balance-moi \textcolor{orange}{Number One} de \textcolor{orange}{Lightspeed Champion} & fr & fr & en \\
\small je peux entendre \textcolor{orange}{Takaj Zhizn} de \textcolor{orange}{Amsterdam Klezmer Band} & fr & fr & nl \\
\small hey mets \textcolor{orange}{Sleepy Shores} de \textcolor{orange}{Laurindo Almeida} & fr & fr & pt \\
\bottomrule
\end{tabular}
\end{table*}

\subsection*{Latency Evaluation Setup}
To evaluate inference efficiency, we measured the average latency for each model over 200 independent single-inference runs. For transformer-based models like XLM-LID and PolyLingua, the measurements were conducted on an A10 GPU. For Sonnet 3.5, latency was calculated using Amazon Bedrock API calls made from an EC2 instance. For lightweight baselines such as FastText and Langdetect, latency was computed by their python package directly on the same EC2 instance. All measurements reflect end-to-end latency per input utterance.

\subsection*{Augmentation}

\section*{Appendix B: Prompt for Sonnet 3.5}\label{appendix: prompt}
We employ the following  prompt to assess the language identification capabilities of proprietary models, like Claude Sonnet 3.5. By specifically telling the model to disregard named entities like artist or brand names, the prompt aims to isolate the linguistic structure of each utterance. To guarantee deterministic results, we set the temperature to 0.0 and use the Bedrock API to invoke the model. The model is expected to return only the ISO 639-1 code of the primary language enclosed in \texttt{<language>...</language>} tags.

\begin{tcolorbox}[colback=gray!5!white, colframe=black!75!black, title=Prompt for Language Detection (Claude Sonnet 3.5)]
Identify any named entities (titles, brand names, people, etc.).\
Ignore these named entities when determining the overall language.\
Focus on the remaining words and grammar.\
DO NOT OUTPUT ANY REASONING. Just output the ISO 639-1 code of the text's primary language between \texttt{<language>...</language>} tags.\

Example:\
Reproduciendo "My Ghoulish Figure" de Conan O'Brien Needs A Friend en Amazon Music.\
\texttt{<language>es</language>}\

\texttt{<text>{utterances}</text>}
\end{tcolorbox}

\section*{Appendix C: Datasets Statistics}

\subsection*{Amazon Massive}
In this section, we visualize the distribution of in-domain language labels and the number of words per utterance. The Amazon MASSIVE dataset exhibits a diverse range of languages and utterance lengths, which presents challenges for consistent language identification.

\begin{figure}[h]
    \centering
    \begin{subfigure}[t]{0.48\textwidth}
        \centering
        \includegraphics[width=\linewidth]{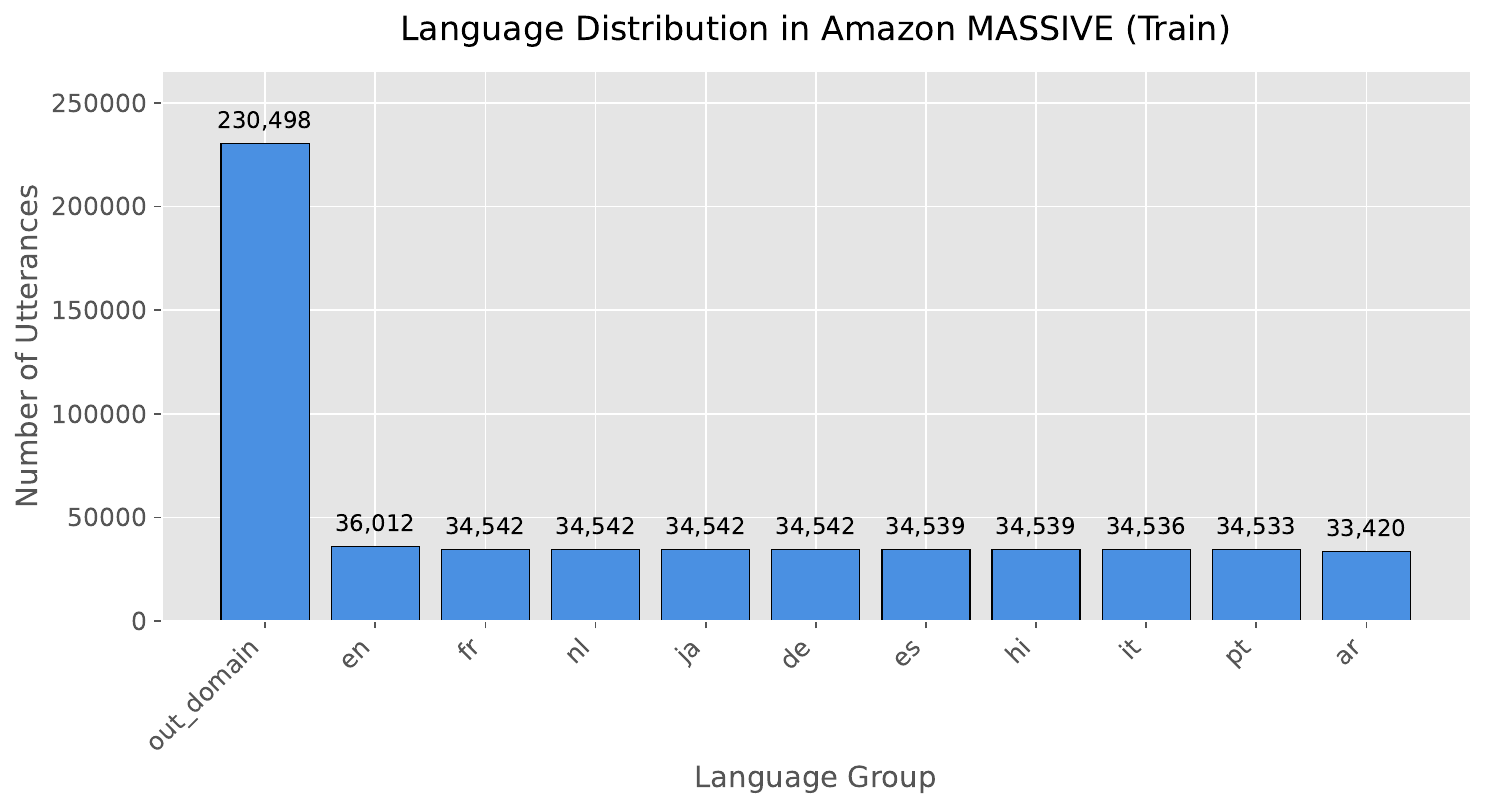}
        \caption{Train set language distribution.}
        \label{fig:train-lang-dist}
    \end{subfigure}
    \hfill
    \begin{subfigure}[t]{0.48\textwidth}
        \centering
        \includegraphics[width=\linewidth]{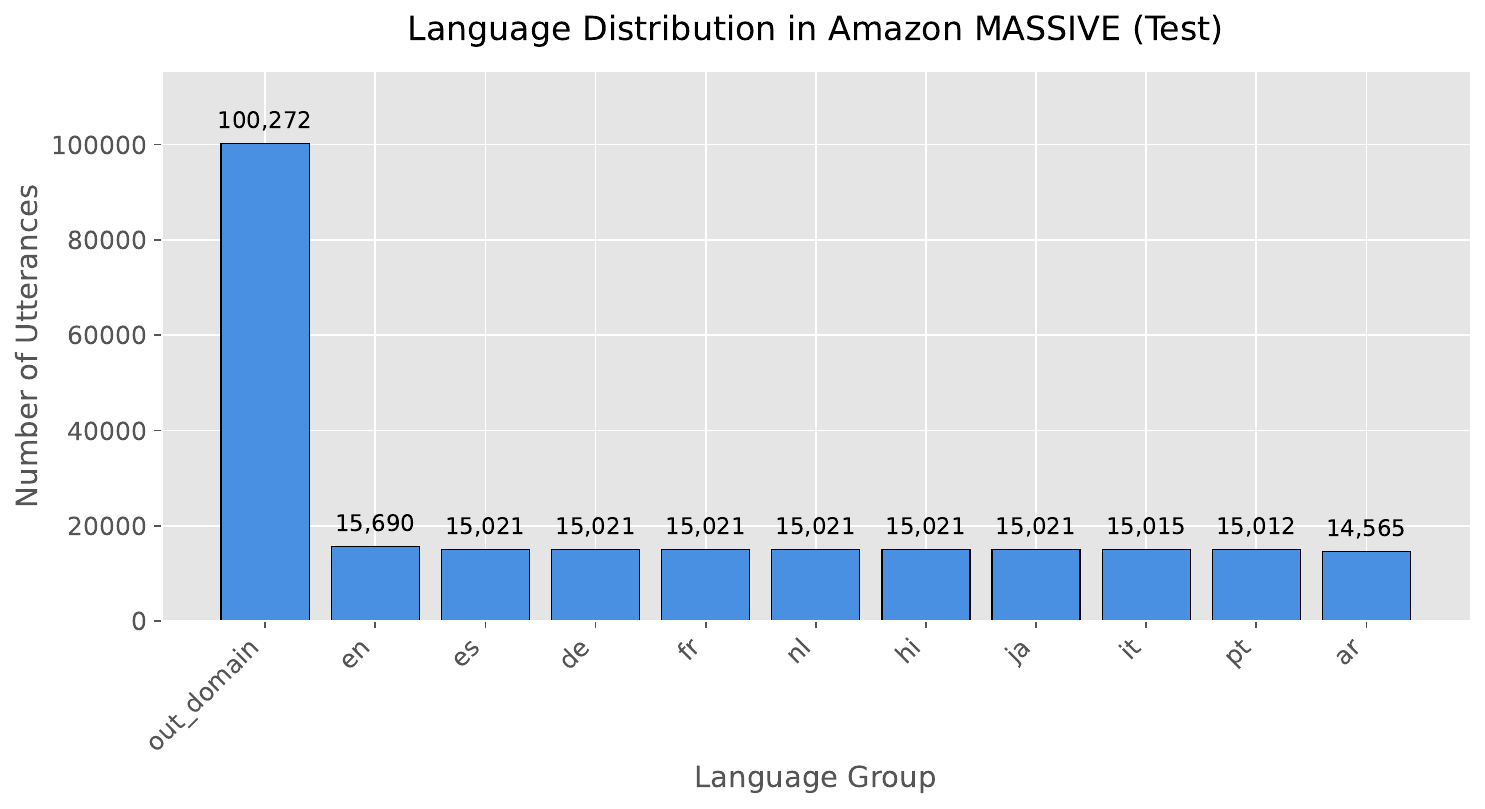}
        \caption{Test set language distribution.}
        \label{fig:test-lang-dist}
    \end{subfigure}
    \caption{Distribution of language labels in the Amazon MASSIVE dataset. We consider ten languages as in-domain; all others are grouped under \texttt{out\_domain}.}
    \label{fig:lang-dist-train-test}
\end{figure}

\begin{figure}[h]
    \centering
    \includegraphics[width=0.75\linewidth]{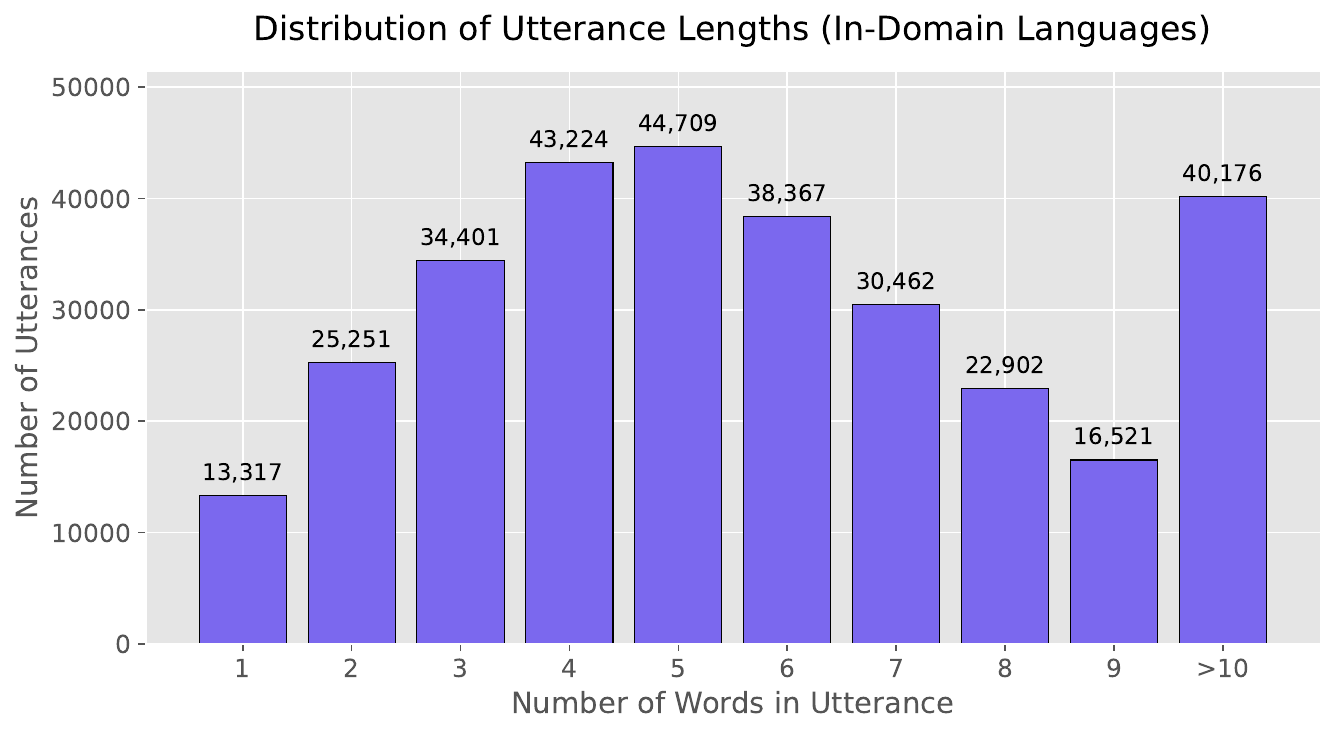}
    \caption{Distribution of utterance lengths (in words) for in-domain languages. Most utterances contain fewer than 7 words, emphasizing the need for robust modeling of short inputs.}
    \label{fig:word-length-dist}
\end{figure}

\begin{table*}[htbp]
  \centering
  \begin{tcolorbox}[
      colback=gray!5!white,
      colframe=black!75!black,
      title={\bf Samples from Song Dataset},
      fonttitle=\bfseries,
      left=2mm,right=2mm,top=2mm,bottom=2mm,
      boxrule=0.5pt,
      arc=2pt,
      enhanced,
    ]
    \small                         
    \begin{tabular}{%
        >{\raggedright\arraybackslash}p{0.68\linewidth} 
        >{\raggedright\arraybackslash}p{0.20\linewidth} 
      }
      \toprule
      \textbf{Utterance} & \textbf{Label} \\
      \midrule
      ich will Nautical Spirits\_ Welcome to the Aquarium von Our Brother The Native hören & de \\
      lass Penny Lane von Orlando Pops Orchestra laufen & de \\
      oh spiel von Whole Wheat Bread & de \\
      kannst du das Lied My Girlhood Among The Outlaws von Maria McKee spielen & de \\
      could you put on Something Right by Julia Fordham & en \\
      play This Room by Mikroboy feat Get Well Soon please & en \\
      i want to listen to Cry now by Elliott & en \\
      would you play Was wir alleine nicht schaffen by Xavier Naidoo & en \\
      pon por favor Life In The Air Age de Be Bop Deluxe & es \\
      tócanos When The Light Came Around de Florian Horwath & es \\
      ponme por favor Fruitcake de Eraserheads & es \\
      este pon People come people go de David Guetta Joachim Garraud Chris Willis & es \\
      mets la chanson Cuando Te Vi de Vicentico & fr \\
      j'aimerais écouter Oh mama oh papa de Ottavo Padiglione & fr \\
      on écoute Third From The Sun de Chrome & fr \\
      est‑ce que tu pourrais mettre LÉternel féminin de Juliette Gréco & fr \\
      mi fai sentire Mannequins di 108 & it \\
      \bottomrule
    \end{tabular}
    \captionof{table}{Some typical samples from the Song dataset}
  \end{tcolorbox}
  \label{tab:song-dataset}
\end{table*}


\end{document}